\documentclass[letterpaper]{article} 
\usepackage[preprint]{aaai2027}  
\usepackage[hyphens]{url}
\usepackage{multirow}
\usepackage{amssymb}
\usepackage{amsmath}
\usepackage{graphicx} 
\usepackage{tikz}
\usetikzlibrary{shapes.geometric, positioning}
\usetikzlibrary{positioning, arrows.meta, calc}
\usepackage{natbib}  
\usepackage{caption} 
\usepackage{algorithm}
\usepackage{algorithmic}

\usepackage{newfloat}
\usepackage{listings}
\usepackage{comment}
\DeclareCaptionStyle{ruled}{labelfont=normalfont,labelsep=colon,strut=off} 
\floatstyle{ruled}
\newfloat{listing}{tb}{lst}{}
\floatname{listing}{Listing}
\usepackage{listings}

\usepackage{booktabs}
\usetikzlibrary{shapes.multipart,shapes.symbols}

\title{Explaining AI Agents Through Execution Traces}

\author {
    Vittoria Vineis\textsuperscript{\rm 1}\thanks{Corresponding author: vineis@diag.uniroma1.it. \\In this preprint version, the source code and trace dataset are  are available upon request.},
    Fabiano Veglianti\textsuperscript{\rm 1},
    Lorenzo Antonelli\textsuperscript{\rm 1},
    Claudia Di Carlo\textsuperscript{\rm 1},
    Matteo Silvestri\textsuperscript{\rm 1},
    Gabriele Tolomei\textsuperscript{\rm 1}
}
\affiliations {
    \textsuperscript{\rm 1}Sapienza University of Rome, Rome, Italy
}

\begin{document}

\maketitle

\begin{abstract}

AI Agents are increasingly deployed in real-world settings, where they interact with external tools and make sequential decisions with limited human oversight. 
This creates a pressing need for reliable and auditable explanations of what an agent did and why. However, traditional Explainable AI (XAI) methods fall short of providing the process-level transparency required for such interactive, multi-step systems, motivating a paradigm shift toward approaches specifically designed for AI Agents.
To address this gap, we present a post-hoc XAI framework that transforms a lengthy agent's execution trace into a structured report and a faithful natural-language explanation explicitly grounded in its observable behavior.
Because it relies solely on execution traces, the framework applies across different agent architectures, environments, and tasks. Human and automated evaluations across multiple benchmarks and architectures show that our framework produces high-quality, trace-faithful explanations while reliably identifying unsupported claims, unjustified actions, and evidence gaps, outperforming naive LLM-generated explanations.

\end{abstract}

\section{Introduction}
Recent advances in Agentic AI are reshaping the capabilities of AI systems by enabling autonomous goal pursuit, adaptive decision-making, and complex task execution with minimal human oversight \cite{acharya2025agentic}. Unlike traditional AI models, AI Agents\footnote{In this paper, we use the terms \emph{AI agents} and \emph{agentic AI systems} interchangeably to refer primarily to LLM-based agents capable of autonomous planning and tool use. While some of the challenges and solutions discussed may also apply to other types of autonomous AI systems, such as agents based on traditional reinforcement learning, our focus is on tool-using LLM agents, which currently constitute the dominant paradigm of Agentic AI.} interact with external tools, maintain internal state, coordinate multiple actions, and adapt their behavior over extended execution horizons. While these capabilities unlock significant opportunities across critical domains, they also introduce new challenges for transparency, accountability, and governance \cite{zhu2026interpreting, shah2026characterizing}, making surface-level transparency no longer sufficient to support user trust or regulatory compliance \cite{khalid2026agentic}. Recent studies further highlight novel failure modes, including error cascades, responsibility gaps, flawed execution monitoring, and multi-step error propagation that obscures the origins of system behavior and outcomes \cite{zhu2026interpreting, shah2026characterizing}, highlighting the need for a rethinking of current XAI practices.

Existing explainability approaches, in fact, were primarily designed for static, single-step predictive models and are increasingly inadequate for explaining the behavior of autonomous agents operating in dynamic environments \cite{chakrabarty2025causal, raza2026transparency, chaduvula2026features}. This challenge is further amplified by the interaction paradigm of Agentic AI, which introduces a complex triadic relationship among users, autonomous agents, and external digital resources, requiring explanation mechanisms that account for multi-layered interactions across interconnected systems \cite{ehsan2026human}. For these reasons, researchers have called for a shift from traditional feature-level explanations toward system-level interpretability capable of capturing agent trajectories, reasoning processes, tool interactions, and execution histories \cite{zhu2026interpreting, raza2026transparency, collaco2026explainable}.

Emerging work advocates for explainability frameworks that treat transparency as a continuous lifecycle property and capable of answering not only what decisions were made, but also why, how, when, and by whom they were influenced \cite{raza2026transparency}. Similarly, other studies warn against the illusion of explainability that Chain-of-Thought disclosures may create, highlighting the need for human-centered approaches that support meaningful understanding rather than merely exposing internal reasoning traces \cite{ehsan2026human}. However, despite the growing recognition of these challenges, research at the intersection of XAI and Agentic AI remains relatively narrow in scope and is largely industry-driven, with a primary focus on tasks such as failure detection and debugging. As a result, existing approaches often fall short of addressing the broader interpretability and accountability debt incurred by these systems across diverse stakeholder groups, including regulators, deployers, and end users.
\\
For these reasons, we introduce a post-hoc, log-grounded framework for generating faithful and auditable structured and natural-language explanations of AI agents' decisions. Our approach leverages execution traces, which are often too lengthy and complex to be readily inspected or interpreted, particularly by non-technical users, to reconstruct and expose the specific sequences of actions and environmental observations that shaped an agent's behavioral trajectory and ultimately led to a given outcome. Because the framework operates exclusively on the logged interaction trace, it can be seamlessly applied across different underlying agent architectures, tool ecosystems, and task domains, enabling broad applicability to real-world AI systems.\\ 
Overall, our work makes the following contributions:
\begin{enumerate}
\item \textbf{A framework for end-user explainability of AI agents.} We introduce a framework specifically designed to make AI agent behavior transparent and understandable to end users. To the best of our knowledge, existing approaches primarily target developer- and system-oriented objectives, such as debugging, monitoring, and failure analysis, rather than treating user-facing explanations as a primary objective. Our framework addresses this gap by defining a process for transforming agent execution traces into faithful and human-readable explanations.

\item \textbf{Two complementary trace-grounded XAI artifacts.} We introduce two complementary artifacts for explaining agent behavior: (i) a structured report that provides a systematic, step-by-step representation of the agent's behavioral trajectory, and (ii) a concise natural-language narrative that communicates the relevant behavior to end users while remaining grounded in the execution trace. The structured report serves as an intermediate, auditable layer connecting the trace to the final explanation and enables precise identification of mismatches between the agent's available knowledge and its observed behavior.

\item \textbf{A taxonomy of agent behavior grounded in knowledge alignment.} We develop a taxonomy that characterizes agent behavior according to the relationship between the agent's available knowledge and information acquired through interaction with the environment. The taxonomy captures aligned and mismatched knowledge states, providing a principled basis for identifying and interpreting different forms of agent behavior from execution traces.

\item \textbf{An empirical evaluation of explanation quality.} We evaluate the framework across two established agentic benchmarks, three agent architectures, and two baseline approaches to generating XAI artifacts. We assess explanations along four dimensions (faithfulness, exhaustiveness, linguistic quality, and overall satisfaction) using both human evaluation and LLM-as-a-judge. Our results show that the proposed method produces structured reports that are substantially more faithful to the execution trace than those obtained by directly prompting an LLM to generate them. Furthermore, for both human and LLM evaluators, narratives grounded in our reports are judged to be more faithful and exhaustive, and receive higher overall ratings, than narratives generated by prompting an LLM to explain the execution trace.
\end{enumerate}

\section{Related Work}
While a growing body of work explores how agentic systems can themselves support explainability tasks \citep{he2026agentic,mehta2026matrag,vineis2026ponte}, to the best of our knowledge no prior work has proposed a framework for making the decision process of Agentic AI systems transparent to end users. Instead, existing work has primarily focused on trace-based analysis for debugging, evaluation and failure diagnosis. Nevertheless, this line of research provides valuable insights that can inform execution trace-grounded explainability methods. In this context, \citet{mazhar2026trace} show that outcome-only evaluation systematically overlooks contamination-induced failures that are only visible through trace inspection. Similarly, \citet{deshpande2025trail}, while introducing a formal taxonomy of agent failures and human-annotated traces (TRAIL benchmark), show that state-of-the-art LLMs remain unreliable at automatically identifying and diagnosing such failures. Addressing procedural compliance in enterprise settings, \citet{sharma2026willful} propose a method (AgentPex) which employs an LLM-as-a-judge to verify whether agent executions satisfy safety and formatting requirements at scale. Likewise, \citet{mulian2026agentfixer} introduce AgentFixer, which analyzes execution logs to detect parsing errors, schema violations, and reasoning inconsistencies before generating corrective actions. Moving beyond LLM-centric approaches, \citet{wang2026agenttrace} reconstructing causal graphs from execution logs to localize failure root causes. Although these methods leverage execution traces as we do, they are primarily designed as developer-facing analysis tools rather than explanation systems. Their objective is to identify why an agent failed, not to communicate how it reached a decision in a faithful and user-understandable manner. By contrast, our framework treats execution traces as the basis for generating trace-grounded explanations that support end-user understanding while simultaneously providing a structured basis for inspecting and auditing agent behavior. This perspective is further supported by \citet{hu2026responsible}, who argue that explicit execution provenance, rather than increasingly sophisticated benchmarks, is a key missing infrastructure for responsibility attribution in Agentic AI. However, their approach is validated only on a proof-of-concept system and does not provide a general framework for generating user-facing explanations, a gap that our work aims to address.

\begin{figure*}[t]
\centering
\resizebox{\textwidth}{!}{%
\begin{tikzpicture}[
    node distance=1.5cm and 1cm,
    every node/.style={
        rectangle, 
        rounded corners, 
        draw=black, 
        fill=white, 
        inner sep=1ex, 
        font=\sffamily, 
        align=left
    },
    scroll/.style={
        draw, 
        inner sep=2ex, 
        shape=tape,
        tape bend top=none, 
        tape bend bottom=wavy, 
        minimum height=2cm, 
        minimum width=2cm, 
        align=center,
        fill=black!8
    },
    largeproc/.style={
        minimum height=4.5cm,
        minimum width=6cm,
        fill=white
    },
    arrow/.style={
        ->, 
        >=stealth, 
        thick, 
        draw=black!70
    }
]

\node (rawexec) [scroll] {Raw\\Execution\\Trace};

\node (proc) [right=of rawexec, largeproc] {
    \textbf{Processes}\\[1ex]
    1. Normalize $\rightarrow \tau$ \\
    2. Decompose $\rightarrow \mathcal{R}$ \\
    3. Extract $\rightarrow \beta, \epsilon^{\text{int}}, \epsilon^{\text{ext}}$ \\
    4. Validate \\
    5. Reconcile $\rightarrow \delta, \gamma$
};

\node (struct) [right=of proc, align=center, fill=blue!12] {
    \textbf{Structured report}\\[0.5ex]
    (first XAI artifact)\\[1.5ex]
    $X(\tau) = \{ \mathcal{G}, n, (a_t, O_t, \Sigma_t, \delta_t, \Gamma_t) \}_{t=0}^T$
};

\node (narr) [right=of struct, minimum height=1.5cm, minimum width=2.5cm, align=center] {Narrator $\psi$};

\node (expl) [right=of narr, align=center, fill=violet!12] {
    \textbf{Natural-language explanation}\\[0.5ex]
    (second XAI artifact)\\[1.5ex]
    $E_1, \dots, E_T, \bar{E}$
};

\draw [arrow] (rawexec.east) -- (proc.west);
\draw [arrow] (proc.east) -- (struct.west);
\draw [arrow] (struct.east) -- (narr.west);
\draw [arrow] (narr.east) -- (expl.west);

\end{tikzpicture}%
}
\caption{Framework overview. The raw AI agent execution trace is processed through a sequential operational pipeline that constructs the structured report $\mathcal{X}(\tau)$. Trace normalization maps heterogeneous execution logs to a canonical step sequence, while requirement decomposition constructs the requirement registry $\mathcal{R}$, to decompose user's goal. At each step, an LLM extracts candidate behavioral commitments and evidence with explicit pointers to the original trace; deterministic validation filters unsupported or ill-typed updates and enforces the framework's grounding constraints. Deterministic reconciliation then computes the per-step state transitions $\delta_t$ and requirement-level grounding reports $\Gamma_t$, yielding the structured tuples $(\Sigma_t,\delta_t,\Gamma_t)$. Finally, the constrained narrator $\psi$ renders $\mathcal{X}(\tau)$ into per-step explanations $E_1,\ldots,E_T$ and a concise overall summary $\bar{E}$.}
\label{fig:framework}
\end{figure*}
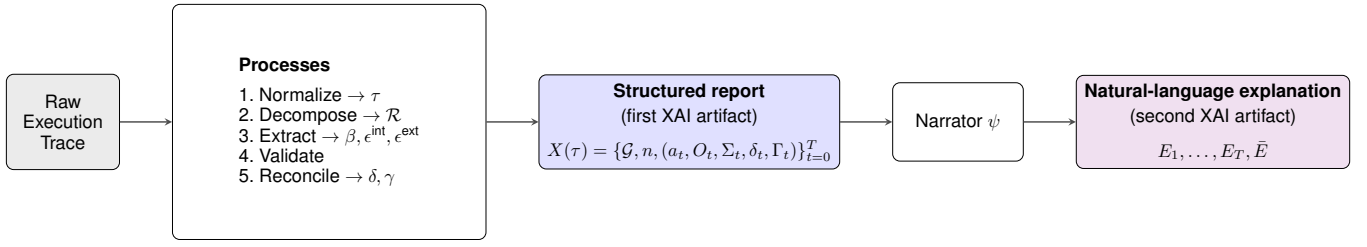

\section{Trace-Grounded XAI Methodology}
\label{sec:framework}
In this section, we formalize the core components of our methodology and their relationships (Section \ref{sec:components}); then we describe their concrete operationalization in a sequential pipeline (Section \ref{sec:pipeline}).

\subsection{Main Components}
\label{sec:components}
As visually summarize in Figure \ref{fig:framework}, the proposed framework transforms a heterogeneous agent execution trace into a structured, trace-grounded representation of the agent's behavioral trajectory, which we refer to as the \emph{structured report} (our first XAI artifact). The report is organized around the goal and its requirements, the actions and observations recorded during execution, and requirement-level states that capture both the agent's behavioral commitments and the evidence available to support them. These components distinguish what the agent appears to commit to from what can be established from the execution trace, while preserving the temporal structure needed to reconstruct how requirements evolve throughout the trajectory. The resulting representation provides the basis for both \emph{natural-language explanations} (our second XAI artifact) and grounding diagnostics.

\paragraph{Goal and Requirements.}
Consider a tool-using LLM agent tasked with achieving a goal $G$, specified by a user's query. We assume that $G$ can be decomposed into a set of atomic, independently verifiable requirements
\(
\mathcal{R}={r_1,r_2,\ldots,r_N},
\)
where each requirement $r_i$ represents a minimal criterion that must be satisfied for the goal to be considered complete. The agent pursues $G$ through iterative interactions with external tools and the environment.

\paragraph{Objective.}
Our objective is to generate a \emph{post-hoc}, \emph{trace-faithful} explanation that makes the agent's reconstructed behavioral trajectory transparent and inspectable, including to non-technical users. Specifically, we produce two complementary XAI artifacts: (i) a structured report that organizes the trajectory into a sequence of epistemically relevant actions and observations, enabling the characterization of transitions in the agent's knowledge--action state and step-by-step inspection of its progress toward the goal; and (ii) a concise natural-language explanation that summarizes this trajectory in human-readable form.

Crucially, the framework does not seek to assess the objective correctness of the agent's final answer, but rather to make transparent the process that led to it. Failure detection is likewise not its primary objective. Although the resulting artifacts can help localize potential errors and characterize inconsistencies exhibited by the agent, these signals concern mismatches between the agent's behavior and the evidence available in the trace. They do not, by themselves, establish failures in the underlying tools or the incorrectness of the agent's knowledge.

\paragraph{Event Trace.}
The fundamental input to our framework is an agent-specific execution trace, generated through the agent's interaction with its environment. This raw trace records agent actions, tool invocations and outputs, together with heterogeneous execution metadata, including timestamps, call identifiers, intermediate states, and framework-specific information. While the structure and granularity of such traces vary across agent implementations, we assume that the recorded information is sufficient to reconstruct the underlying behavioral trajectory. We denote the resulting epistemically meaningful \emph{event trace} by
\(
\tau=(e_1,e_2,\ldots,e_T),
\)
where each event $e_t$ captures an action, observation, or other interaction that may alter the information available to the agent at that point in the trajectory. By convention, $T$ denotes the terminal step at which the agent produces
its final response to the user; all preceding steps $t<T$ correspond to
intermediate actions and observations along the trajectory. This representation abstracts away implementation-specific logging details while preserving the information required to characterize how the agent progresses from the initial query toward its goal.

\paragraph{Knowledge--Action State Transitions.}
The event trace $\tau$ provides the basis for constructing a structured representation of the agent's behavioral trajectory. We denote by $a_t$ the identifier of the agent action at step $t$ and by $o_{t,k}$ the identifier of the $k$-th observation associated with that step. At each step $t$, we reconstruct a requirement-indexed knowledge--action state \[
\Sigma_t =
\left(
\mathcal{R}_t,
\left\{
\beta_t(r), \epsilon_t(r)
\right\}_{r \in \mathcal{R}_t}
\right),
\]
where $\mathcal{R}_t \subseteq \mathcal{R}$ is the set of requirements introduced up to step $t$, $\beta_t(r)$ records the agent's behavioral commitment with respect to requirement $r$, and $\epsilon_t(r)$ records the evidence available for that requirement in the execution trace.
For each $r \in \mathcal{R}_t$, $\beta_t(r)$ ans $\epsilon_t(r)$ take values in $\{\texttt{open}, \texttt{satisfied}\}$.

The behavioral component $\beta_t$ is a trace-grounded reconstruction of the commitment expressed by the agent's subsequent behavior. In particular, its value is inferred from the next relevant agent action:
\(
\beta_t(r)=f_\beta(a_{t+1},r),
\)
where $f_\beta$ determines whether the action selected after step $t$ indicates that the agent treats requirement $r$ as resolved. 

The evidence component is instead decomposed into two provenance-specific channels:
\[
\epsilon_t(r)=
\bigl(
\epsilon_t^{\mathrm{int}}(r),
\epsilon_t^{\mathrm{ext}}(r)
\bigr),
\]
whose statuses may independently be $\texttt{open}$ or $\texttt{satisfied}$ and may therefore disagree for the same requirement $r$ at a given step $t$.
The internal component is supported by observations corresponding to the agent's own surfaced information, such as reflections or tool inputs, whereas the external component is supported by observations returned by the environment, such as tool outputs and execution feedback. 

On both behavioral and evidence axes, the \texttt{satisfied} status requires an explicit
supporting reference in the trace.  Formally, $\forall\, r \in \mathcal{R}_t^*$ and $\forall x \in \{\beta\} \cup \{\epsilon^{c} : c \in \{\mathrm{int},\mathrm{ext}\}\}$,
\[
x_t(r) = \texttt{satisfied} \Longrightarrow \operatorname{supp}_{x}(r) \neq \varnothing.
\]

with $supp_{\beta}(r)$ pointing to the action $a_{t+1}$ (or the goal $G$)
whose content expresses the commitment, and $supp_{\epsilon^{c}}(r)$
pointing to at least one observation $o_{t,k}$ of the corresponding provenance
whose content entails $r$. When no admissible reference exists, the
status remains \texttt{open}. This invariant makes every positive
assertion in $\Sigma_t$ pointer-cited and directly auditable against the
raw trace.

As the agent proceeds through the observed action--observation sequence, its $\Sigma_t$ states evolve to reflect changes in requirements, behavioral commitments, and available evidence. We summarize each consecutive state change with a transition operator
\[
\delta_t \;=\; \Delta(\Sigma_t,\Sigma_{t+1}),
\]
which identifies changes in requirement status across $\beta$ and $\epsilon$, including requirements becoming satisfied or reopening, as well as the introduction of new requirements.

\paragraph{Behaviour-Evidence Discrepancies.} Given the status of $\beta$ and $\epsilon$ for each requirement $r \in \mathcal{R}_t$, we deterministically assign a grounding label based on the joint configuration of the agent's behavioral commitment and the available internal and external evidence:
\[
\gamma_t(r)\;=\;
f\!\left(
\beta_t(r),\,
\epsilon_t^{\mathrm{int}}(r),\,
\epsilon_t^{\mathrm{ext}}(r)
\right).
\]

The label characterizes whether the agent's behavioral commitment is
consistent with the evidence available in the trace and, when it is not,
identifies the nature of the discrepancy.
The complete $2^3$ configuration space is reported in Table~\ref{tab:diagnostic_conditions}. We organize these configurations into two groups. \emph{Concordant} configurations are those in which the behavioral commitment is consistent with the available evidence. This group includes both cases in which a requirement is supported by internal and/or external evidence and the \emph{genuine gap}, in which the requirement remains unresolved and no supporting evidence is available. \emph{Mismatches} are configurations in which the behavioral commitment diverges from the available evidence. The four mismatch categories constitute the core diagnostic output of $\gamma$.
Aggregating $\gamma_t$ over all requirements yields the per-step
grounding report
\[
  \Gamma_t \;=\;
  \bigl\{\,\bigl(r,\gamma_t(r)\bigr)\,:\,
    r\in\mathcal{R}_t,\;\gamma_t(r)\in\mathcal{L}\,\bigr\},
\]
where $\mathcal{L}$ is the label set of
Table~\ref{tab:diagnostic_conditions}. By construction
$\Gamma_t=\varnothing$ when every requirement is warranted, so
$\Gamma_t$ acts as the step-level record of the discrepancies between
behavioral commitment and trace evidence.

\begin{table*}[t]
\centering
\small
\setlength{\tabcolsep}{6pt}
\renewcommand{\arraystretch}{1.15}
\begin{tabular}{ccc ll}
\toprule
$\beta$ & $\epsilon^{int}$ & $\epsilon^{ext}$ & Label & Interpretation \\
\midrule
\multicolumn{5}{l}{\emph{Concordant configurations (bahaviour matches available evidence)}}\\
\midrule
1 & 1 & 1 & \textit{Fully Grounded}      & Resolved with both internal and external evidence.\\
1 & 1 & 0 & \textit{Self Grounded}       & Resolved from internal evidence only.\\
1 & 0 & 1 & \textit{Externally Grounded}     & Resolved from external evidence only.\\
0 & 0 & 0 & \textit{Genuine Gap}               & Unresolved and no supporting evidence available.\\
\midrule
\multicolumn{5}{l}{\emph{Discordant configurations (behaviour and evidence disagree)}}\\
\midrule
1 & 0 & 0 & \textit{Ungrounded}         & Resolved without trace-grounded evidence.\\
0 & 1 & 0 & \textit{Missed Internal Evidence} & Internal evidence is available but not reflected in behaviour.\\
0 & 0 & 1 & \textit{Missed External Evidence}       & External evidence is available but not reflected in behaviour.\\
0 & 1 & 1 & \textit{Missed Recognition}        & Both evidence sources are available but not reflected in behaviour.\\
\bottomrule
\end{tabular}
\caption{Diagnostic configurations derived from behaviour--evidence
reconciliation. Here, $\beta$, $\epsilon^{int}$, and $\epsilon^{ext}$ denote behavioral committment, internal
evidence, and external evidence, respectively, measured with respect to a specific  requirement $r$ at step $t$;  each equals $1$ when the
corresponding state is $\texttt{satisfied}$.}
\label{tab:diagnostic_conditions}
\end{table*}

\paragraph{Structured report.} Aggregating across the trajectory, the framework's first XAI artifact is
the \emph{structured report}
\[
  \mathcal{X}(\tau) \;=\;
  \Bigl(\,G,\;\bigl\{\bigl(a_t,\,\mathcal{O}_t,\,\Sigma_t,\,\delta_t,\,\Gamma_t\bigr)
                    \bigr\}_{t=0}^{T}\,\Bigr),
\]
where $\mathcal{O}_t=\{o_{t,k}\}_{k=1}^{K_t}$ denotes the set of observations
associated with step $t$. 

The structured report records, for every step,
the action taken, the observations received, the reconstructed
knowledge--action state, the transition summary, and the grounding report
(with $\delta_0$ omitted and $\delta_T$ defined against the terminal
state). $\mathcal{X}(\tau)$ is directly inspectable and provides the
audited input to the natural-language explanation described next.

\paragraph{Final Narrative.} The second XAI artifact is then produced by a constrained narrator
\[
  \psi:\;\mathcal{X}(\tau)\;\longrightarrow\;
        \bigl(E_1,\,E_2,\,\ldots,\,E_T,\,\bar{E}\bigr),
\]
where each per-step explanation
\[E_t=\psi_{\mathrm{step}}(a_t,\mathcal{O}_t,\Sigma_t,\delta_t,\Gamma_t)\]
verbalizes the epistemic change at step $t$ and any divergence recorded
in $\Gamma_t$, and $\bar{E}$ compresses the trajectory into an
episode-level summary. \\

\subsection{Operational Pipeline}
\label{sec:pipeline}

Below, we describe the concrete instantiation of the framework components introduced above and their organization into a sequential operational pipeline.

\paragraph{Trace Normalization.}
The raw execution trace is mapped to the canonical step sequence required
by the framework. Each step preserves the action, its observations, their
semantic types, and the terminal condition. The representation uses a
fixed observation vocabulary---user query, tool input, tool output,
reflection, error, and final response---and separates tool inputs from
tool outputs so that agent-emitted information remains distinguishable
from environment-returned information. The resulting canonical sequence
is the common interface to all subsequent stages, decoupling the analysis
from the underlying agent architecture and enabling the same pipeline to
be applied across heterogeneous paradigms. 

\paragraph{Requirement Decomposition.}
The normalized goal is decomposed into the requirement registry
$\mathcal{R}$ using a summarize-then-decompose strategy inspired from \citet{cho2026ask}. The
decomposition targets atomicity at the level required for independent
status assessment: requirements concerning different entities are kept
distinct whenever their satisfaction can be established independently.
Each requirement is assigned a stable identifier and classified as either
a \emph{content requirement} (information that must appear in the answer)
or a \emph{constraint requirement} (formatting or structural condition).
Depending on the application, requirements may be generated upfront,
incrementally, or through a hybrid strategy that admits newly discovered
subgoals during execution.

\paragraph{Local Behaviour--Evidence Extraction.}
At each step, an LLM produces local updates to the requirement-level
state. Given the information available at step $t$ together with the
subsequent action $a_{t+1}$, the extractor proposes behavioral commitments
and evidence items, each accompanied by pointers to the specific trace
elements that support them. We recall that behavioural commitment is inferred from
$a_{t+1}$; evidence proposals are assigned to the internal or external
channel according to the provenance of their cited observations.

\paragraph{Validation and Grounding.}
Candidate updates are subjected to deterministic validation. Behavioral
supports must correspond to the action designated as the behavioral basis,
and evidence supports must reference observations present in the trace and
admissible for their assigned channel; ill-typed or unsupported references
are discarded. When entailment verification is enabled, an evidence
channel is marked \texttt{satisfied} only if the cited content
additionally entails the corresponding requirement. Validated updates are
then incorporated into the state subject to the framework's consistency
and monotonicity constraints.

\paragraph{Deterministic Reconciliation.}
The validated state sequence is processed deterministically to obtain the
per-step summaries. Consecutive states yield the transition $\delta_t$,
and the joint status of behavioral commitment and the two evidence
channels determines the label $\gamma_t(r)$ for every requirement, which
aggregate into the step-level report $\Gamma_t$. The resulting
$(\Sigma_t,\delta_t,\Gamma_t)$ tuples complete the structured report
$\mathcal{X}(\tau)$.

\paragraph{Narrative Generation.}
The narrator $\psi$ is instantiated as a
constrained LLM call over $\mathcal{X}(\tau)$. Because every belief,
evidence item, and diagnostic has already been grounded in the trace,
narrative generation reduces to verbalizing an audited execution history
rather than reconstructing reasoning from raw logs. The narrator is
constrained to reference only elements of its input---requirement
identifiers, action identifiers $a_t$, observation identifiers $o_{t,k}$,
and their statuses---so that every generated statement remains
trace-grounded and no content is introduced beyond $\mathcal{X}(\tau)$.
The resulting explanation reports what the agent attempted, what
information it acquired, which requirements became satisfied, and where
potential failures or inconsistencies occurred.
\section{Experiments}
We evaluate the framework through a controlled study to examine the reliability and trace-grounded fidelity of the structured report and the quality of the natural-language explanations it supports. The following sections describe the experimental setup (Section \ref{sec:exp_setup}) and evaluation protocols (Section \ref{sec:evaluation}).
\begin{table*}[t]
\centering
\small
\setlength{\tabcolsep}{6pt}
\renewcommand{\arraystretch}{1.15}
\begin{tabular}{ll r cc ccc}
\toprule
& & &
\multicolumn{2}{c}{Structured report\,---\,failures\,(\%)} &
\multicolumn{3}{c}{Narrative\,---\,failures\,(\%)} \\
\cmidrule(lr){4-5}\cmidrule(l){6-8}
Benchmark & Agent & \# traces &
  \textsc{Our} & \textsc{LLM} &
  \textsc{Our--Our} & \textsc{LLM--Our} & \textsc{LLM-only} \\
\midrule
GAIA      & Plan-and-act & 116 & 0 & 1\,(0.9\%)  & 0 & --- & 25\,(21.6\%) \\
GAIA      & ReAct        & 116 & 0 & 9\,(7.8\%)  & 0 & --- & 40\,(34.5\%) \\
GAIA      & TRAIL        & 116 & 0 & 3\,(2.6\%)  & 0 & --- & 28\,(24.1\%) \\
HotpotQA  & Plan-and-act & 100 & 0 & $1\,(1.0\%)$ & 0 & --- & $5\,(5.0\%)$   \\
HotpotQA  & ReAct        & 100 & 0 & 0  & 0 & --- & $1\,(1.0\%)$   \\
\midrule
\textbf{Total} & & \textbf{548} &
  \textbf{0} & \textbf{14\,(2.6\%)} &
  \textbf{0} & --- & \textbf{99\,(18.1\%)} \\
\bottomrule
\end{tabular}
\caption{Evaluation corpus and pipeline reliability. For each of the five
(benchmark $\times$ agent) subsets we report the number of traces
processed and the number of \emph{failed} generations per pipeline
configuration (percentages relative to the row total). The
\textit{Structured report} block covers the intermediate representation
($\mathcal{X}(\tau)$), produced either by our deterministic pipeline
(\textsc{Our}) or by a single-shot LLM prompt with the same schema
(\textsc{LLM}). The \textit{Narrative} block covers the final
natural-language explanation ($\bar{E}$), produced by: (i)\textsc{Our--Our}, our
narrator applied to our structured report (the full pipeline described in Section \ref{sec:pipeline});
(ii) \textsc{LLM-Our}, our narrator applied to an directly LLM-generated version of the structured report; and (iii) \textsc{LLM-only},  an LLM prompted to summarise the raw
trace directly in natural language.}
\label{tab:corpus}
\end{table*}

\subsection{Experimental Setup}
\label{sec:exp_setup}
We evaluate the proposed framework on execution traces from two question-answering (QA) benchmarks and an external trace corpus. The evaluation is designed along two complementary dimensions. First, we assess the contribution of the proposed structured analysis by comparing our XAI artifacts with outputs generated directly by an LLM under matched input and model conditions. Second, we evaluate the quality of the resulting artifacts using both human assessment and LLM-as-a-judge evaluation. This design allows us to disentangle the contribution of the structured, trace-grounded analysis from that of the underlying language model used for generation.
\paragraph{Trace Corpus.}
We focus our empirical evaluation on tool-empowered QA tasks because they provide a controlled setting for assessing explanations of agent behavior. QA benchmarks typically specify well-defined user goals and information-seeking objectives, making requirement satisfaction more amenable to systematic assessment and reducing reliance on subjective judgments of task success. Moreover, tool-mediated QA produces explicit action--observation sequences, enabling us to assess whether explanations faithfully account for observable agent behavior rather than merely whether the final answer is correct. This makes QA well suited to evaluating transparency and auditability while enabling larger-scale evaluation at lower annotation cost than tasks requiring extensive expert assessment. The choice of QA is specific to our empirical evaluation, not a limitation of the framework, which can be applied to other agentic tasks given suitable execution traces.

Among existing QA resources, TRAIL~\citep{deshpande2025trail} is, to our knowledge, the only dataset providing publicly available execution traces for a QA benchmark. We therefore use TRAIL as our external trace source and build our evaluation around its underlying GAIA~\citep{mialon2024gaia} question set. The original TRAIL agent and its execution environment are not available to us, and the released trajectories do not expose explicit external tool calls, primarily capturing the agent's internal reasoning. We therefore complement TRAIL with independently generated trajectories exhibiting richer and more heterogeneous tool use. Specifically, we implement two additional agent configurations and execute them on the same GAIA questions, enabling controlled comparison across different agent architectures and execution structures while preserving a common task set.

The first agent is a \emph{ReAct} agent \citep{yao2022react}, which interleaves reasoning and tool use within a single execution loop. The second is a \emph{plan-and-act} agent that follows an iterative plan--act--reflect loop with an explicit step for routing between actions. Both agents use \textit{gemini-3.1-flash-lite}\footnote{\url{https://deepmind.google/models/model-cards/gemini-3-1-flash-lite/}}as the underlying LLM and share a heterogeneous tool set comprising Wikipedia, Tavily web search, calculator, Python code execution, PDF reader, and text editor. This tool set covers information retrieval, web browsing, numerical computation, programmatic analysis, and document processing, yielding diverse tool interactions representative of practical agentic QA systems.

For direct comparison with the external trace source, we use the $116$ GAIA questions for which TRAIL provides executions and run both independently implemented agents on the same questions. We additionally evaluate the two agents on $100$ HotpotQA~\citep{yang2018hotpotqa} questions sampled from the distractor validation split. This complementary multi-hop QA setting allows us to assess whether the framework generalizes beyond the GAIA/TRAIL task distribution.

The resulting evaluation corpus comprises five benchmark--agent subsets: GAIA--plan-and-act ($116$ traces), GAIA--ReAct ($116$), GAIA--TRAIL ($116$), HotpotQA--plan-and-act ($100$), and HotpotQA--ReAct ($100$), for a total of $548$ execution traces.
\paragraph{XAI Artifact Generation.}
Each trace is processed independently using the pipeline described in Section \ref{sec:pipeline}. The LLM-based extraction and validation stage uses \textit{gemini-3.1-flash-lite}, while state updates, reconciliation, and report construction follow the deterministic procedures defined by the framework. Requirements are generated in upfront mode, such that the complete requirement registry is constructed from the user goal at $t=0$. We evaluate both artifacts produced by the framework: the structured report $\mathcal{X}(\tau)$ and its natural-language rendering. Across the $548$ execution traces, we obtain $1{,}096$ structured reports from the two structured-report configurations and $1{,}644$ natural-language explanations from the three narrative configurations.

\paragraph{Baselines and Ablations.}
We consider two complementary comparisons to isolate the contributions of the structured representation and its trace-grounded construction. To distinguish the source of each artifact, we use paired notation $(S,N)$, where $S$ denotes the source of the structured report and $N$ the source of the narrative.

The \textsc{Our--Our} configuration denotes the complete proposed pipeline: the structured report is constructed by our trace-grounded pipeline and the resulting natural language explanation is rendered by our constrained narrator $\psi$.

The \textsc{LLM--Our} configuration isolates the contribution of the structured-report construction procedure. The LLM receives the same canonical trace representation, requirement vocabulary, and target schema used by our framework, but is prompted to generate the complete structured report autonomously. This preserves the conceptual representation introduced by our method while bypassing its multi-stage extraction, validation, state-update, and deterministic reconciliation procedures. The resulting report is then passed unchanged to the same constrained narrator $\psi$. This comparison tests whether the benefits of our structured representation can be obtained simply by prompting an LLM to populate the proposed schema, or instead depend on the trace-grounded construction process.

Finally, we use the \textsc{LLM-only} notation to denote direct narrative generation without an intermediate structured representation, which correspond to use the LLM directly as the explainer. In this case the original trace is provided to a single LLM call, which is prompted to generate the natural-language explanation directly. The prompt is designed based on a review of existing approaches to LLM-based generation and evaluation of XAI narratives \cite{silvestri2025survey}, with the goal of providing a competitive baseline informed by the state of the art. This configuration therefore provides an end-to-end comparison in which the model must derive and verbalize the explanation directly from the execution trace.

All configurations use \textit{gemini-3.1-flash-lite} for LLM-based generation, including the construction of intermediate structured reports and final narratives. Where comparisons are available, configurations operate on the same input trace and use the same underlying model, varying only the representation and generation procedure. This controls for differences in model capacity and focuses the comparison on the contribution of the proposed structured and trace-grounded pipeline.

\paragraph{Generation Failures.}
Original execution traces can be long and structurally complex, making single-shot generation particularly demanding when the model must process the complete trajectory and produce the XAI artifacts. We therefore allow up to $20$ generation attempts for each requested artifact. A generation is considered \emph{failed} if the requested artifact has not been successfully produced after these $20$ attempts\footnote{For \textsc{LLM-only} configuration, narrative generation may fail for excessively long inputs, as the trace and system prompt alone can exceed 250k tokens, thereby reaching the context-window limit of the free Gemini plan used for generating XAI outputs across all configurations.}. Failed generations are excluded from downstream artifact-quality evaluation but are retained in the reliability analysis reported in Table \ref{tab:corpus}. As shown in the table, failures occur primarily for the single-shot LLM configurations and are more frequent for the longer and more complex traces (TRAIL has shorter execution trace than the one obtained from our agentic configurations). Our proposed pipeline records no failures, owing to its design choices described in Section \ref{sec:framework}, which make the framework robust to long and structurally complex execution traces.

\subsection{Evaluation}
\label{sec:evaluation}

We evaluate the framework at two levels: the fidelity of the intermediate
structured representation and the quality of the generated narratives. The
evaluation uses two complementary protocols. First, on a stratified subset of
the execution corpus, human annotators independently reconstruct the
structured representation directly from the original execution traces. These
human-annotated representations provide a trace-grounded reference for
evaluating the structured reports produced by our framework and by the LLM
baseline. Second, we evaluate the generated narratives using both human
assessment and an LLM-as-a-judge protocol~\cite{zheng2023judgingllmasajudgemtbenchchatbot},using \emph{Muse-Spark-1.2-Contributor} as evaluation model.

\paragraph{Human evaluation subset and reference annotation.}
We conduct human evaluation on a stratified subset of the execution corpus,
comprising $32$ traces from $12$ queries across three agent systems
(ReAct, plan-and-act, and TRAIL agent). The ReAct and
plan-and-act agents contribute $12$ traces each, covering $9$ GAIA and
$3$ HotpotQA questions, while TRAIL contributes $8$ GAIA traces\footnote{One sampled
GAIA question has no corresponding TRAIL execution and is therefore excluded
from the TRAIL subset.}. GAIA questions are selected through stratified random
sampling over difficulty and number of tool interactions, following the
characterization provided by the original GAIA benchmark paper. For HotpotQA, whose original difficulty characterization is based on query length and is not meaningful to our setting, we assign each query a difficulty level and tool-use stratum
using the same criteria adopted for GAIA.

For each selected trace, a human annotator independently constructs a
structured representation directly from the original execution trace,
following the representation schema defined in
Section \ref{sec:framework}. This human-annotated representation serves as
the reference for evaluating the two structured-report variants: the report
produced by our full pipeline and the report obtained by prompting the LLM to
instantiate the same schema. Because the reference is constructed directly
from the trace and independently of both generated reports, it provides an
independent basis for assessing the fidelity of the intermediate
representation.

Considering the two structured-output methods (\textsc{Our} and
\textsc{LLM}) and the three narrative-generation settings
(\textsc{Our-Our}, \textsc{LLM-Our}, and \textsc{LLM-only}), and excluding
cases in which artifact generation failed (see Table \ref{tab:corpus}), the human evaluation covers
$63$ structured reports and $89$ narratives.

\begin{table*}[t]
\centering
\small
\setlength{\tabcolsep}{5pt}
\renewcommand{\arraystretch}{1.3} 
\caption{Narrative evaluation criteria and evaluation configurations.}
\label{tab:narrative_metrics}
\begin{tabular}{p{0.18\textwidth} p{0.33\textwidth} p{0.33\textwidth} p{0.08\textwidth}}
\toprule
\textbf{Metric} & \textbf{Definition} & \textbf{Evaluation configuration} & \textbf{Scale} \\
\midrule
\multirow{3}{=}{\textbf{M1 -- Faithfulness}}
& \multirow{3}{=}{The narrative accurately reflects the relevant behavior represented by its reference, without unsupported claims, omissions, or semantic distortions.}
& \textbf{M1-a} -- Source report: Human + LLM judge
& 3-point \\
& & \textbf{M1-b} -- Trace level: Human judge
& 3-point \\
& & \textbf{M1-a\_our} -- Our report: LLM judge
& 3-point \\
\midrule

\textbf{M2 -- Exhaustiveness} (Content)
& The narrative contains all essential information needed to understand the agent's behavior and the relevant steps leading to its outcome.
& Human + LLM judge
& 4-point \\
\midrule

\textbf{M3 -- Linguistic Quality} (Delivery)
& The narrative is clear, logically organized, concise, communicatively efficient, and easy to understand, without unnecessary or unsupported verbosity.
& Human + LLM judge
& 4-point \\
\midrule

\textbf{M4 -- Overall Quality} (Holistic)
& Overall, the narrative provides an appropriate and satisfactory explanation of the agent's execution, considering both its content and presentation.
& Human + LLM judge
& 4-point \\

\bottomrule

\end{tabular}
\end{table*}
\paragraph{Structured Report Evaluation}
On the human-annotated subset, we assess the faithfulness of the two structured-report variants.  Because this comparison requires reliable assessment of the correspondence between the structured representation and the underlying execution, and LLM-as-a-judge methods cannot be assumed to provide faithful judgments, particularly for lengthy and complex execution traces, we restrict this evaluation to the human-annotated subset. This provides a direct assessment of whether the intermediate representation accurately captures the behavior exhibited in the trace.

We assess the fidelity of each generated structured report to the human-annotated reference obtain by the original execution trace using an ad hoc metric we called the \emph{Normalized
Faithfulness of the Intermediate representation} (NFI). For a generated
report, let $R_t^{GT}$ denote the set of ground-truth requirements associated
with step $t$ and $R_t^{gen}$ the corresponding requirements in the predicted
report. We define

\begin{equation}
\mathrm{NFI}
=
\frac{
\sum_t
\mathrm{matched}\!\left(R_t^{gen},R_t^{GT}\right)
}{
\sum_t |R_t^{GT}|
},
\label{eq:nfi}
\end{equation}

where $\mathrm{matched}(\cdot,\cdot)$ counts the ground-truth requirements
correctly represented by the generated report. The denominator is therefore
the total number of ground-truth requirements across all steps, yielding a
score in $[0,1]$. NFI directly measures how much of the trace-grounded
behavior captured by the human reference is recovered by each structured
representation. To adopt a maximally strict criterion, we count as an error any discrepancy at a given step, including the omission of a ground-truth requirement, a missed $\delta_t$ or $\gamma_t$, or an incorrect classification of $\beta_t$ or $\epsilon_t$, including an incorrect supporting evidence assignment. 


\begin{table*}[t]
\centering
\small
\caption{NFI: mean and standard deviation by method and agent, overall and by number-of-steps group.}
\label{tab:nfi_method_agent_stepgroup}
\begin{tabular}{llrrrrrrrrrrrr}
\toprule
 &  & \multicolumn{3}{c}{Overall} & \multicolumn{3}{c}{$\leq 10$} & \multicolumn{3}{c}{11--30} & \multicolumn{3}{c}{$\geq 31$} \\
\cmidrule(lr){3-5} \cmidrule(lr){6-8} \cmidrule(lr){9-11} \cmidrule(lr){12-14}
METHOD & AGENT & $n$ & mean & std & $n$ & mean & std & $n$ & mean & std & $n$ & mean & std \\
\midrule
OUR & ReAct & 12 & 0.904 & 0.092 & 5 & 0.902 & 0.099 & 3 & 0.860 & 0.098 & 4 & 0.940 & 0.090 \\
OUR & Plan-and-act & 12 & 0.842 & 0.212 & 8 & 0.794 & 0.237 & 4 & 0.938 & 0.125 & -- & -- & -- \\
OUR & TRAIL & 8 & 0.644 & 0.282 & 8 & 0.644 & 0.282 & -- & -- & -- & -- & -- & -- \\
LLM & ReAct & 11 & 0.075 & 0.131 & 5 & 0.116 & 0.161 & 3 & 0.083 & 0.144 & 3 & 0.000 & 0.000 \\
LLM & Plan-and-act & 12 & 0.057 & 0.149 & 8 & 0.062 & 0.177 & 4 & 0.045 & 0.090 & -- & -- & -- \\
LLM & TRAIL & 8 & 0.046 & 0.077 & 8 & 0.046 & 0.077 & -- & -- & -- & -- & -- & -- \\
\bottomrule
\end{tabular}
\end{table*}

\paragraph{Narrative Evaluation.}
Following consolidated practices identified in the narrative XAI literature \cite{silvestri2025survey}, we evaluate natural language explanations along four dimensions, summarized in Table \ref{tab:narrative_metrics}.

\emph{Faithfulness} is assessed at two main levels: narrative-to-report faithfulness (M1-a) and trace-level faithfulness (M1-b). The former measures whether the narrative accurately reflects the behavior represented in the structured report from which it was generated, without introducing unsupported claims, omitting relevant information, or altering the meaning of the represented behavior. The narrative-to-source-report assessment applies to the \textsc{Our-Our} and \textsc{LLM-Our} narratives, for which the structured report is available. 
Trace-level faithfulness measures whether the narrative faithfully reflects the behavior observed in the original execution trace. This assessment requires comparison against the human-annotated ground truth and is therefore performed only through human evaluation on the human-evaluated subset, providing a direct assessment of trace-level faithfulness.
As a further specification of narrative-to-report faithfulness, in the LLM-as-a-judge evaluation we also assess each narrative against the structured report produced by our framework (M1-a\_our).  This configuration is used for cross-validation: we compare the resulting LLM-judge scores with the corresponding human assessments of the narrative's trace-level faithfulness to the original execution trace. Agreement between the two provides evidence that the structured representation produced by our framework preserves the behaviour relevant to narrative faithfulness and can therefore serve as a suitable proxy for trace-level evaluation. This comparison also provides evidence that LLM-as-a-judge evaluation can capture, at least partially, trace-level faithfulness when the judge has access to the structured representation rather than the original execution trace.  

\emph{Exhaustiveness} (M2) measures whether the narrative covers all the
essential information required to explain the agent's behavior. \emph{Linguistic quality} (M3) assesses clarity, logical organization,
conciseness, and the absence of unnecessary or unsupported verbosity. \emph{Overall quality} (M4) captures the evaluator's holistic assessment
of the narrative as an explanation of the agent's execution.

To ensure consistent application of the evaluation criteria, we operationalize each metric through an explicit evaluation rubric, which specifies the criteria and corresponding rating anchors for both human and LLM-judge assessment. Faithfulness is evaluated on a three-point ordinal scale (“Totally agree,” “Partially agree,” “Disagree”), whereas M2–M4 are evaluated on a four-point Likert scale (“Strongly agree,” “Agree,” “Disagree,” “Strongly disagree”).

\paragraph{Agreement analysis.}
Before using LLM-as-a-judge for scalable evaluation, we assess its alignment with human judgments, on the common scored subset. For each criterion rated by both evaluators, we report raw percentage agreement after collapsing the ordinal scale into positive and negative judgments, together with quadratic-weighted Cohen's $\kappa$ ($\kappa_w$), which preserves the ordinal structure of the ratings and penalizes larger disagreements more strongly than near-misses.

Both measures are interpreted in conjunction with the marginal distribution of the ratings. In particular, when narratives from stronger configurations receive positive judgments almost uniformly, the negative class becomes small and chance-corrected agreement can be attenuated: high raw agreement may therefore coexist with a low $\kappa_w$ when little disagreement remains for chance correction to capture. We therefore additionally report, for each criterion, the proportion of positive ratings assigned by each evaluator and the judge's \emph{negative recall}, defined as the proportion of cases rated negatively by humans that are also rated negatively by the judge. Negative recall is particularly relevant for scalable evaluation, as it measures whether the judge identifies the failures that human evaluators would flag, independently of a shared tendency to assign positive ratings.

Statistics are pooled across the three narrative configuration. They are computed over the paired ratings available for each criterion; consequently, the effective sample size varies across criteria because of missing or inapplicable annotations.

\begin{table*}[t]
\centering
\small
\caption{Agreement between the human annotator and the LLM judge, pooled over conditions (bold) and split by condition. H/J = percentage of ratings on the agree side for the human and judge, respectively; Agr.\ = raw agreement on the agree/disagree collapse; $\kappa_w$ = quadratic-weighted Cohen's $\kappa$; Neg.\ rec.\ = share of human-negative ratings also rated negative by the judge, with counts in parentheses. $^{\dagger}$ marks rows with fewer than three human-negative ratings, for which $\kappa_w$ is not informative; raw agreement and negative recall should be preferred. M1-a\_our vs M1-b is a cross-level comparison: the judge scores the narrative against our structured report, whereas the human scores it against the original trace.}
\label{tab:alignment}
\begin{tabular}{llrrrr}
\toprule
Measure & Condition ($n$) & H/J (\%) & Agr. (\%) & $\kappa_w$ & Neg.\ rec. \\
\midrule

\multirow{3}{*}{\textbf{M1-a}}
& \textbf{All (63)}  & \textbf{95/98} & \textbf{94} & \textbf{-0.09} & \textbf{0\% (0/3)} \\
& Our--Our (32)       & 100/97 & 97 & $-0.06^{\dagger}$ & n/a \\
& LLM--Our (31)       & 90/100 & 90 & $-0.05$ & 0\% (0/3) \\

\addlinespace

\multirow{4}{*}{\textbf{M1-a\_our vs M1-b}}
& \textbf{All (86)}  & \textbf{81/60} & \textbf{74} & \textbf{0.63} & \textbf{88\% (14/16)} \\
& Our--Our (32)       & 100/97 & 97 & $0.09^{\dagger}$ & n/a \\
& LLM--Our (31)       & 87/48 & 61 & 0.46 & 100\% (4/4) \\
& LLM-only (23)       & 48/26 & 61 & 0.45 & 83\% (10/12) \\

\addlinespace

\multirow{4}{*}{\textbf{M2}}
& \textbf{All (86)}  & \textbf{71/87} & \textbf{79} & \textbf{0.28} & \textbf{36\% (9/25)} \\
& Our--Our (32)       & 94/94 & 94 & $0.22^{\dagger}$ & 50\% (1/2) \\
& LLM--Our (31)       & 48/71 & 71 & 0.36 & 50\% (8/16) \\
& LLM-only (23)       & 70/100 & 70 & 0.00 & 0\% (0/7) \\

\addlinespace

\multirow{4}{*}{\textbf{M3}}
& \textbf{All (86)}  & \textbf{81/86} & \textbf{84} & \textbf{0.52} & \textbf{44\% (7/16)} \\
& Our--Our (32)       & 94/100 & 94 & $0.27^{\dagger}$ & 0\% (0/2) \\
& LLM--Our (31)       & 87/100 & 87 & 0.16 & 0\% (0/4) \\
& LLM-only (23)       & 57/48 & 65 & 0.40 & 70\% (7/10) \\

\addlinespace

\multirow{4}{*}{\textbf{M4}}
& \textbf{All (86)}  & \textbf{62/80} & \textbf{67} & \textbf{0.41} & \textbf{33\% (11/33)} \\
& Our--Our (32)       & 97/91 & 94 & $0.13^{\dagger}$ & 100\% (1/1) \\
& LLM--Our (31)       & 55/68 & 68 & 0.48 & 50\% (7/14) \\
& LLM-only (23)       & 22/83 & 30 & -0.06 & 17\% (3/18) \\

\bottomrule
\end{tabular}
\end{table*}

\begin{table*}[t]
\centering
\small
\caption{Hypothesis tests on the LLM-as-judge corpus and comparison with human-evaluation effects. Inferential statistics are reported for the LLM-as-judge corpus, while human-evaluation effects are reported for descriptive comparison only.}
\label{tab:hypotheses_judge}
\begin{tabular}{llrrlrrrrrl}
\toprule
& &
\multicolumn{2}{c}{Human} &
\multicolumn{7}{c}{LLM judge} \\
\cmidrule(lr){3-4}
\cmidrule(lr){5-11}
H & Measure &
$n$ & $\Delta$ [95\% CI] &
$n$ & $W$ & $T$ & $L$ & $\Delta$ [95\% CI] & $p$ & $p_{\mathrm{BH}}$ \\
\midrule
H1 & M2
& 26 & 0.85 [0.62, 1.07]
& 417 & 18 & 395 & 4 & 0.07 [0.02, 0.12] & 0.0031 & 0.0031 \\

H1 & M3
& 26 & 1.04 [0.65, 1.36]
& 418 & 353 & 65 & 0 & 1.18 [1.12, 1.25] & 0.0001 & 0.0001 \\

H1 & \textbf{M4}
& 26 & 1.35 [1.00, 1.64]
& 416 & 206 & 209 & 1 & 0.62 [0.55, 0.69] & 0.0001 & 0.0001 \\

H2 & M2
& 31 & 1.03 [0.61, 1.45]
& 500 & 124 & 373 & 3 & 0.49 [0.41, 0.58] & 0.0001 & 0.0001 \\

H2 & M3
& 31 & 0.39 [-0.03, 0.82]
& 500 & 30 & 438 & 32 & -0.00 [-0.03, 0.03] & 0.6418 & 0.8022 \\

H2 & \textbf{M4}
& 31 & 1.00 [0.58, 1.42]
& 497 & 186 & 305 & 6 & 0.82 [0.72, 0.93] & 0.0001 & 0.0001 \\

\bottomrule
\end{tabular}
\end{table*}

\begin{table}[t]
\centering
\small
\caption{H3 direction consistency for the LLM-as-judge corpus. Cell = mean paired difference within that agent architecture.}
\label{tab:h3_consistency_judge}
\begin{tabular}{llrrr}
\toprule
H & Measure & ReAct & Plan-and-act & TRAIL \\
\midrule
H1 & M2 & +0.08 & +0.03 & +0.15 \\
H1 & M3 & +1.18 & +1.11 & +1.37 \\
H1 & M4 & +0.63 & +0.55 & +0.77 \\
H2 & M2 & +0.45 & +0.59 & +0.39 \\
H2 & M3 & -0.01 & -0.01 & +0.03 \\
H2 & M4 & +0.76 & +0.93 & +0.73 \\
\bottomrule
\end{tabular}
\end{table}

\paragraph{Hypotheses and analysis protocol.}
We formulate four directional hypotheses prior to analysis, corresponding to the principal claims tested in the evaluation:

\begin{itemize}
\item \textbf{H0 -- Structured-representation fidelity.}
Our structured report is more faithful and complete as a step-by-step representation of the execution trace than the freely LLM-generated report, as measured by the normalized faithfulness index (NFI) against human-annotated ground truth (M0). H0 provides a further basis for using our structured report as the reference for evaluating narrative faithfulness in M1-a\_our.

\item \textbf{H1 -- End-to-end superiority.}
Narratives produced by the full pipeline, \textsc{Our-Our}, receive higher ratings than narratives generated by directly prompting an LLM to explain the execution trace, \textsc{LLM-only}, across all narrative-quality criteria: trace-level faithfulness (M1-b), exhaustiveness (M2), linguistic quality (M3), and overall satisfaction (M4).

\item \textbf{H2 -- Value of the structured representation.}
Holding the narrator fixed, narratives produced from our structured report, \textsc{Our-Our}, receive higher ratings than those produced from an LLM-generated structured report, \textsc{LLM-Our}, on structured-report faithfulness (M1-a), trace-level faithfulness (M1-b), exhaustiveness (M2), linguistic quality (M3), and overall satisfaction (M4).

\item \textbf{H3 -- Invariance across agent architectures.}
The direction of the H1 and H2 effects is preserved across the three source agent architectures, such that the observed advantages of \textsc{Our-Our} over \textsc{LLM-Only} and \textsc{Our-Our} over \textsc{LLM-Our} do not depend on the architecture generating the execution trace.
\end{itemize}

All comparisons are paired on the same execution traces. Because the human-evaluated subset comprises 12 queries, with each query executed by up to three agents, ratings associated with the same query are not treated as independent observations. For each directional comparison, we therefore obtain $p$-values from a one-sided paired permutation test in which the sign of the paired difference is flipped at the query level and applied jointly to all agents associated with that query. With 12 queries, the resulting null distribution contains $2^{12}$ possible sign assignments and can be enumerated exhaustively, yielding exact $p$-values. We control multiplicity using the Benjamini--Hochberg procedure across criteria within each hypothesis and designate M4 (overall satisfaction) as the primary confirmatory outcome, with the remaining criteria treated as secondary. Each comparison is reported together with the mean paired difference, a cluster bootstrap interval over queries, and the corresponding win, tie, and loss counts.
For analysis, ordinal categories are mapped to consecutive integers in
ascending order ($1$--$3$ for faithfulness, $1$--$4$ for M2--M4), and $\Delta$
denotes the mean paired difference on that coding; it should be read as an
average movement in rating categories rather than as an interval-scale
quantity. The tests themselves do not depend on this choice: under the null the
sign-flip distribution is exact for any statistic computed from the paired
differences, and the reported win, tie and loss counts are invariant under any
order-preserving recoding of the categories.
Confidence intervals for $\Delta$ are obtained by a nonparametric cluster
bootstrap with $B=5{,}000$ replicates and a fixed seed. The resampling unit is
the query: each replicate draws with replacement as many clusters as the
contrast contains and takes each drawn cluster entire, with all
of its (query, agent) blocks. The number of clusters per replicate is therefore
fixed at $k$, whereas the number of blocks is random with expectation equal to
the observed $n$, since a query contributes between one and three blocks; the
statistic is the unweighted mean over the blocks of a replicate, matching the
point estimate. Reported intervals are the $2.5$th and $97.5$th percentiles of
the $B$ replicate means. The same procedure, with the same $B$ and the same
resampling unit, produces the intervals accompanying the correlations. At $k\approx 12$ the percentile bootstrap is known to
under-cover, so on the human subset these intervals are indicative and no
inference rests on them.

H3 is evaluated through consistency of effect direction across agent architectures; in the current study, this analysis is limited to the human-evaluated subset, while LLM-as-a-judge-based assessment of the remaining architectures is left for future work.

\section{Results}
In the main text, we report the corresponding LLM-as-a-judge analysis, conducted on a larger evaluation corpus and therefore providing greater statistical power. Where relevant, we compare these results with the directional findings observed in the human evaluation. Detailed hypothesis tests and further analyses of the human-evaluation results are reported in Appendix~\ref{app}. 
\paragraph{Structured-representation fidelity (H0).}
Table~\ref{tab:nfi_method_agent_stepgroup} reports NFI for the two structured-report
variants against the human-annotated reference. Our structured reports substantially
outperform those generated by directly prompting the LLM: across 63 reports, our
method achieves $\mathrm{NFI}=0.816$ ($\mathrm{SD}=0.219$), compared with $0.061$
($\mathrm{SD}=0.124$) for the LLM baseline. The advantage is
consistent across all three agent architectures and shows no evident degradation
with trace length, whereas the LLM-generated reports decline as the number of steps
increases: under the strict matching criterion $74\%$ of them score zero, versus none
of ours. These results support H0 and the use of our structured report as the
reference for evaluating narrative faithfulness.

\paragraph{LLM-as-judge validation.}
Before applying the LLM-as-judge evaluation at scale, we assess its agreement with
human annotations on the subset evaluated by both methods (Table~\ref{tab:alignment}).
Raw agreement is highest for \textsc{Our-Our} ($94$--$97\%$), intermediate for
\textsc{LLM-Our} ($61$--$90\%$), and lower for \textsc{LLM-only} (up to $70\%$). These
values must be read together with the marginal distributions: where ratings are highly
skewed---\textsc{Our-Our} has fewer than three human-negative ratings per
criterion---high raw agreement carries little information and $\kappa_w$ is
correspondingly attenuated.

For criteria with sufficient variation, agreement is substantial: pooled $\kappa_w$ is
$0.63$ for cross-level faithfulness, $0.52$ for linguistic quality, and $0.41$ for
overall quality. The main discrepancy sits in the calibration, since the judge is systematically
more positive than humans, most visibly on the weaker configurations (e.g., for M4
the \textsc{LLM-only} positive rate is $22\%$ for humans but $83\%$ for the judge),
which compresses the differences between conditions, particularly for exhaustiveness.
Consequently the judge recovers only $9/25$ human-negative M2 cases and $11/33$ M4
cases, so M2-based judge scores in particular cannot replace human assessment. This
may reflect differing weightings of surface fluency, apparent completeness, and
trace-grounded evidence between human and model-based evaluation.

The cross-level faithfulness comparison (M1-a\_our vs.\ M1-b) provides an additional
test of the structured representation. Without access to the ground-truth trace, the
judge nonetheless identifies $14/16$ narratives that humans judge unfaithful to the
trace. Together with the NFI analysis, this supports using the structured report as a
practical proxy for trace-level faithfulness assessment when direct trace-based
evaluation is costly.
\paragraph{End-to-End Superiority (H1)}

We first evaluate H1, which tests whether the complete \textsc{Our-Our} pipeline yields higher scores than the \textsc{LLM-only} configuration. As shown in Table~\ref{tab:hypotheses_judge}, all three metrics exhibit positive effects on the LLM-as-judge corpus. The effect is smallest for M2 ($\Delta=0.07$), but remains statistically significant. The effects are substantially larger for M3 ($\Delta=1.18$) and M4 ($\Delta=0.62$), both with $p<0.0001$. The corresponding human-evaluation estimates follow the same ordering for M3 and M4, although their magnitudes are larger. Overall, these results support H1: replacing the LLM-only pipeline with the complete proposed pipeline improves performance across all evaluated criteria, with the strongest advantage observed for M3.

\paragraph{Value of the Structured Representation (H2)}

H2 isolates the contribution of the structured representation by comparing configurations that differ in whether this representation is incorporated into the generation process. The results in Table~\ref{tab:hypotheses_judge} reveal a differentiated pattern across metrics. M2 shows a positive effect ($\Delta=0.49$), as does M4, with a substantially larger effect ($\Delta=0.82$); both effects are statistically significant. In contrast, M3 shows essentially no difference between the two configurations ($\Delta\approx0$, $95\%$ CI $[-0.03,0.03]$, $p=0.6418$). This pattern is also reflected in the human-evaluation estimates, although the effects are generally larger in magnitude (see Appendix \ref{app}). This is expected given that M3 evaluates linguistic realization, which depends on the narrator, and the two configurations compared under H2 share the same narrator.

\paragraph{Invariance Across Agent Architectures (H3)}
H3 examines whether the effects observed under H1 and H2 are consistent across agent architectures. Table~\ref{tab:h3_consistency_judge} reports the mean paired differences separately for ReAct, Plan-and-act, and TRAIL agent. Under H1, all three measures show a positive effect across all architectures, yielding consistent directions. The same pattern holds under H2 for M2 and M4, with positive effects across agent configurations. In contrast, M3 remains effectively null under H2 across all three architectures ($\Delta=-0.01$, $-0.01$, and $+0.03$, respectively). This behavior is expected given that the two configurations compared under H2 share the same narrator across architectures, while M3 evaluates linguistic realization produced by that narrator. Thus, changing the underlying agent architecture does not introduce a systematic M3 difference between the two configurations.

\section{Conclusion}

We presented a post-hoc framework for explaining AI agents through their execution traces. By transforming heterogeneous traces into an auditable structured representation that explicitly separates behavioral commitments from available internal and external evidence, the framework provides both step-level diagnostics and trace-grounded natural-language explanations. Across three agent architectures and two QA benchmarks, the structured reports were substantially more faithful to human references than direct LLM generation, while narratives produced from these reports achieved higher faithfulness, exhaustiveness, and overall quality than direct trace-to-text explanations. The results further show that the structured representation can serve as a practical proxy for trace-level faithfulness when direct trace inspection is costly.

Current limitations include the evaluation's focus on tool-mediated QA tasks and a relatively small human-evaluated subset. We stress that, by construction, the framework can explain only behavior and evidence recorded in the execution trace; incomplete or insufficient logging necessarily limits what can be reconstructed.  Future work will evaluate the framework on a wider range of agentic tasks and increasingly complex execution scenarios.

\bibliography{biblio}

\appendix
\section{Detailed Human-Evaluation Results}
\label{app}

\paragraph{End-to-end superiority (H1).}
H1 compares narratives produced by the full pipeline (\textsc{Our-Our}) with those
generated by directly prompting an LLM to explain the execution trace
(\textsc{LLM-only}). The full pipeline significantly outperforms the direct explainer
on all four criteria, with $p_{\mathrm{BH}}=0.0019$ throughout
(Table~\ref{tab:hypotheses_human}). The mean paired differences are $\Delta=1.35$ for overall quality, $1.11$ for trace-level
faithfulness, $1.04$ for linguistic quality, and $0.85$
for exhaustiveness. Because  Across the four criteria and 26 paired traces, \textsc{Our-Our} receives
higher ratings in 80 of 104 comparisons, ties in 23, and loses only once.

\begin{figure*}[t]
    \centering

    \begin{minipage}[t]{0.48\textwidth}
        \centering
        \includegraphics[width=\linewidth]{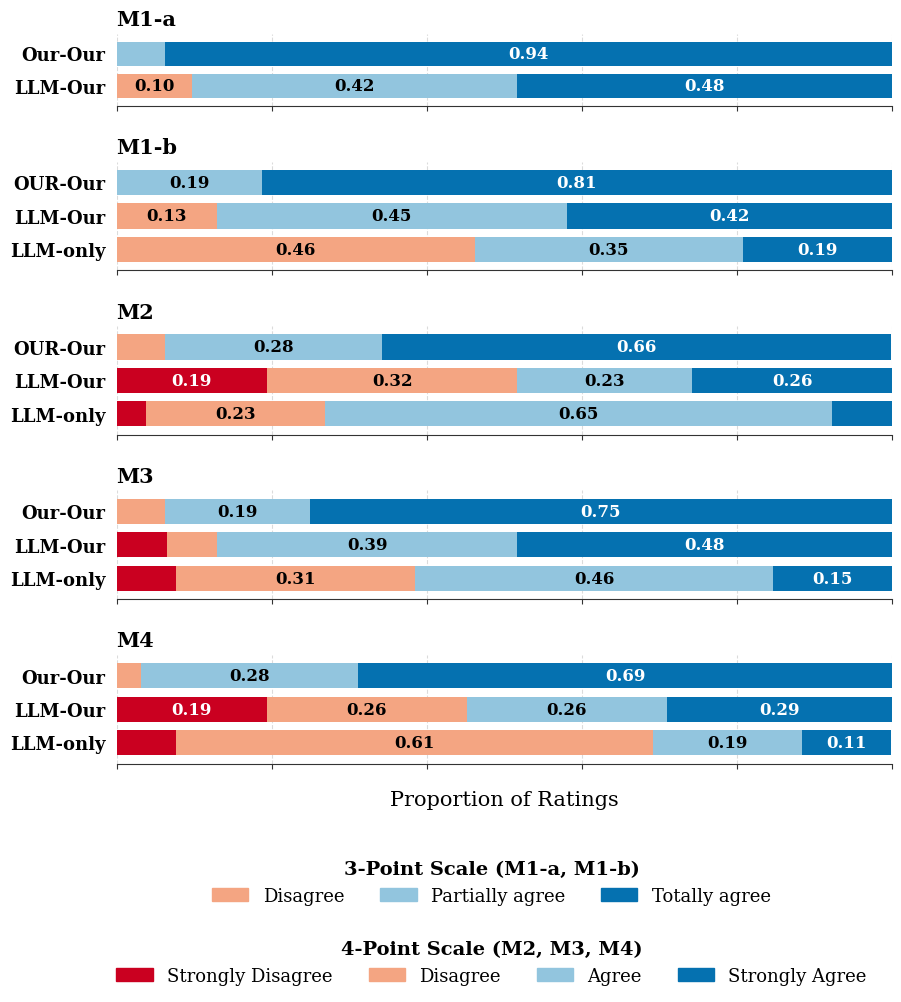}
        \caption{Distribution of Likert ratings provided by human evaluators across the five evaluation dimensions. Bars show the proportion of responses in each category for Our--Our, LLM--Our, and LLM-only. By design, dimension M1-a does not include an LLM-only configuration.}
        \label{fig:human_all}
    \end{minipage}
    \hfill
    \begin{minipage}[t]{0.48\textwidth}
        \centering
        \includegraphics[width=\linewidth]{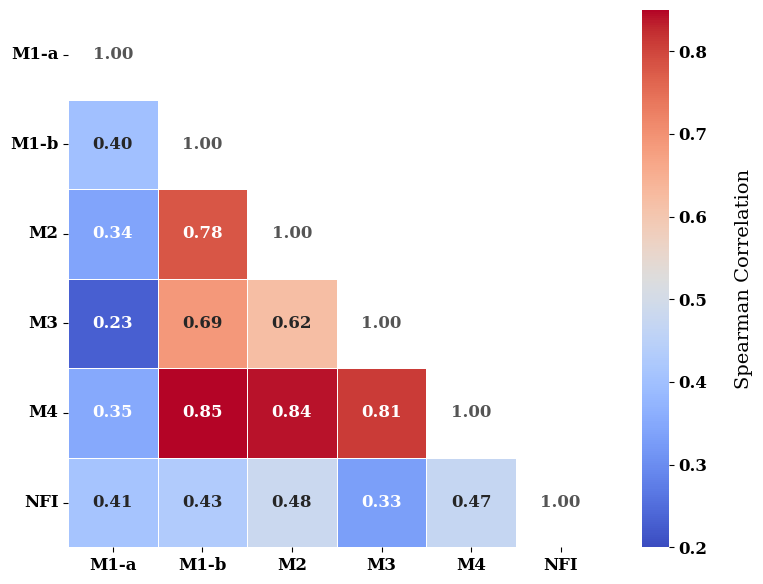}
        \caption{Spearman correlation across evaluation dimensions. Coefficients are computed over pairwise-complete ratings: $n=63$ when at least one dimension in the pair evaluates only structured outputs, and $n=89$ when both dimensions evaluate narratives.}
        \label{fig:spearman_corr}
    \end{minipage}

\end{figure*}

The distribution of ratings (Figure~\ref{fig:human_all}) further clarifies the
difference. For exhaustiveness, \textsc{LLM-only} receives positive ratings in $73\%$
of cases but only $8\%$ reach the highest category. This apparent strength does not
extend to trace-level faithfulness: $46\%$ of \textsc{LLM-only} narratives receive an
outright \emph{Disagree}. Overall quality follows the faithfulness pattern: only
$31\%$ of \textsc{LLM-only} narratives are rated positively, versus $97\%$ for
\textsc{Our-Our}. Thus direct prompting can produce narratives perceived as reasonably
complete and fluent, but these qualities do not translate into faithful explanations
of the trace, consistent with the benefit of grounding generation in the structured,
step-level representation.

\paragraph{Value of the Structured Representation (H2).}
H2 holds the narrator fixed and varies only the structured report, isolating the
contribution of the structuring stage. \textsc{Our-Our} significantly outperforms
\textsc{LLM-Our} on four of five criteria (Table~\ref{tab:hypotheses_human}):
exhaustiveness ($\Delta=1.03$), overall
quality ($\Delta=1.00$), faithfulness to the
source report ($\Delta=0.55$), and
trace-level faithfulness ($\Delta=0.52$).
Linguistic quality does not significantly differ ($\Delta=0.39$), consistent with both configurations sharing the same
narrator. The effect is smaller and more frequently tied than in the end-to-end
comparison, with 75 wins, 70 ties, and 10 losses across 155 paired comparisons.

The distribution of ratings (Figure~\ref{fig:human_all}) shows how the deficit in the
LLM-generated report propagates: narratives built on it remain largely faithful to the
report itself ($90\%$ positive M1-a) but exhibit substantially lower content quality
(exhaustiveness positive in $48\%$ vs.\ $94\%$ for \textsc{Our-Our}; overall quality
$55\%$ vs.\ $97\%$), while linguistic quality stays similar ($87\%$ vs.\ $94\%$). The
primary effect of the structuring stage is thus on the content available to the
narrator rather than on linguistic realization. For M1-a, $17$ of $31$ paired
comparisons are ties while \textsc{Our-Our} wins the remaining $14$ and loses none:
when the LLM report is adequate both configurations produce faithful narratives, and
our pipeline's advantage emerges primarily when the intermediate representation
becomes limiting.

\paragraph{Invariance Across Agent Architectures (H3).}
Table~\ref{tab:h3_consistency} reports the H1 and H2 contrasts separately for each
agent architecture. All subgroup effects are positive: none of the 27
architecture-specific effects reverses direction or is zero, so H3 is supported in
terms of effect-direction consistency. The subgroups contain only 8--12 queries,
insufficient for a reliable condition-by-architecture interaction test, so we report
directions and magnitudes without $p$-values. The confirmatory outcome, overall
quality, is particularly stable (H1: $+1.33$, $+1.33$, $+1.38$; H2: $+1.00$, $+1.08$,
$+0.88$ across ReAct, Plan-and-act, and TRAIL). Effect magnitudes vary for individual
criteria---H2 faithfulness to the source report ranges from $+0.27$ on ReAct to
$+1.12$ on TRAIL, and linguistic quality shows the smallest effects ($+0.64$, $+0.33$,
$+0.12$). At the judge level, the same contrasts stay direction-consistent across
architectures (Table~\ref{tab:h3_consistency_judge}), the one exception being H2 on
M3, which is $\approx 0$ everywhere as expected under a fixed narrator.

\paragraph{Relations among criteria.}
Figure~\ref{fig:spearman_corr} shows strong associations between overall quality (M4) and
trace-level faithfulness ($\rho=0.85$), exhaustiveness ($\rho=0.84$), and delivery
($\rho=0.81$), suggesting that human holistic evaluation reflects substantive content
and faithfulness rather than fluency alone. Faithfulness to the structured report
(M1-a) is comparatively weakly associated with the narrative criteria
($\rho=0.23$--$0.40$): a narrative may faithfully reflect an imperfect report, so M1-a
alone is insufficient to characterize explanation quality. NFI shows moderate
associations ($\rho=0.33$--$0.48$), indicating that the structured representation
constrains narrative quality without fully determining it.
\begin{table*}[t]
\centering
\begin{minipage}[t]{0.60\textwidth}
\centering
\setlength{\tabcolsep}{3pt}
\renewcommand{\arraystretch}{0.85}
\small
\caption{Hypothesis tests on the human-evaluated subset (12 queries $\times$ 3 agents). $n$ = paired blocks; W, T and L = wins, ties and losses for the first condition; $\Delta$ = mean paired difference with a cluster bootstrap CI over queries (indicative at 12 clusters); $p$ = exact one-sided query-level permutation; $p_{\mathrm{BH}}$ = Benjamini--Hochberg corrected within hypothesis. Bold = pre-designated confirmatory outcome. H1: Our--Our $>$ LLM-only; H2: Our--Our $>$ LLM--Our; H0: our structured report $>$ LLM structured report.}
\label{tab:hypotheses_human}
\begin{tabular}{llrrrrlrr}
\toprule
H & Measure & $n$ & W & T & L & $\Delta$ [95\% CI] & $p$ & $p_{\mathrm{BH}}$ \\
\midrule
H1 & M1-b & 26 & 20 & 6 & 0 & 1.11 [0.79, 1.39] & 0.0019 & 0.0019 \\
H1 & M2 & 26 & 18 & 7 & 1 & 0.85 [0.62, 1.07] & 0.0010 & 0.0019 \\
H1 & M3 & 26 & 20 & 6 & 0 & 1.04 [0.65, 1.36] & 0.0019 & 0.0019 \\
H1 & \textbf{M4} & 26 & 22 & 4 & 0 & 1.35 [1.00, 1.64] & 0.0010 & 0.0019 \\
H2 & M1-a & 31 & 14 & 17 & 0 & 0.55 [0.32, 0.79] & 0.0019 & 0.0032 \\
H2 & M1-b & 31 & 13 & 17 & 1 & 0.52 [0.28, 0.71] & 0.0029 & 0.0037 \\
H2 & M2 & 31 & 18 & 10 & 3 & 1.03 [0.61, 1.45] & 0.0015 & 0.0032 \\
H2 & M3 & 31 & 12 & 15 & 4 & 0.39 [-0.03, 0.82] & 0.0781 & 0.0781 \\
H2 & \textbf{M4} & 31 & 18 & 11 & 2 & 1.00 [0.58, 1.42] & 0.0015 & 0.0032 \\
H0 & \textbf{NFI} & 31 & 31 & 0 & 0 & 0.75 [0.69, 0.82] & 0.0002 & 0.0002 \\
\bottomrule
\end{tabular}
\end{minipage}
\hfill
\begin{minipage}[t]{0.35\textwidth}
\centering
\setlength{\tabcolsep}{4pt}
\renewcommand{\arraystretch}{0.9}
\small
\caption{H3 direction consistency, according to human evaluations. Entries are mean paired differences within each agent architecture. H3 is supported when the effect retains its sign across all three architectures.}
\label{tab:h3_consistency}
\begin{tabular}{llrrr}
\toprule
H & Measure & ReAct & Plan-and-act & TRAIL \\
\midrule
H1 & M1-b & +1.44 & +0.78 & +1.12 \\
H1 & M2   & +0.56 & +0.89 & +1.12 \\
H1 & M3   & +1.00 & +0.89 & +1.25 \\
H1 & M4   & +1.33 & +1.33 & +1.38 \\
H2 & M1-a & +0.27 & +0.42 & +1.12 \\
H2 & M1-b & +0.82 & +0.25 & +0.50 \\
H2 & M2   & +1.27 & +0.83 & +1.00 \\
H2 & M3   & +0.64 & +0.33 & +0.12 \\
H2 & M4   & +1.00 & +1.08 & +0.88 \\
\bottomrule
\end{tabular}
\end{minipage}
\end{table*}

\end{document}